\documentclass[10pt,twocolumn,letterpaper]{article}

\usepackage[pagenumbers]{cvpr} % To force page numbers, e.g. for an arXiv version

\definecolor{cvprblue}{rgb}{0.21,0.49,0.74}
\usepackage[pagebackref,breaklinks,colorlinks,allcolors=cvprblue]{hyperref}

\def\paperID{*****} % *** Enter the Paper ID here
\def\confName{CVPR}
\def\confYear{2026}

\title{CtrlAttack: A Unified Attack on World-Model Control in Diffusion Models}

\author{Shuhan Xu\\
Wuhan University\\
\and
Siyuan Liang \\
Nanyang Technological University
\and
Hongling Zheng \\
Wuhan University \\
\and
Yong Luo \\
Wuhan University \\
\and
Han Hu \\
Beijing Institute of Technology \\
\and
Lefei Zhang \\
Wuhan University \\
\and 
Dacheng Tao \\
Nanyang Technological University \\
}

\begin{document}
\maketitle
\begin{abstract}
Diffusion-based image-to-video (I2V) models increasingly exhibit world-model-like properties by implicitly capturing temporal dynamics. However, existing studies have mainly focused on visual quality and controllability, and the robustness of the state transition learned by the model remains understudied.
To fill this gap, we are the first to analyze the vulnerability of I2V models, find that temporal control mechanisms constitute a new attack surface, and reveal the challenge of modeling them uniformly under different attack settings.
Based on this, we propose a trajectory-\textbf{\underline{c}}on\textbf{\underline{tr}}o\textbf{\underline{l}} attack, called CtrlAttack, to interfere with state evolution during the generation process.
Specifically, we represent the perturbation as a low-dimensional velocity field and construct a continuous displacement field via temporal integration, thereby affecting the model's state transitions while maintaining temporal consistency; meanwhile, we map the perturbation to the observation space, making the method applicable to both white-box and black-box attack settings.
% Experimental results show that on 4 models and VIPSeg dataset, even under low-dimensional and strongly regularized perturbation constraints, our method can still significantly disrupt temporal consistency by increasing the [xx] by [xx\%] while keeping the variation of the [xx] within [xx], thus revealing the potential security risk of I2V models at the level of state dynamics.
Experimental results show that even under low-dimensional and strongly regularized perturbation constraints, our method can still significantly disrupt temporal consistency by increasing the attack success rate (ASR) to over 90\% in the white-box setting and over 80\% in the black-box setting, while keeping the variation of the FID and FVD within 6 and 130, respectively, thus revealing the potential security risk of I2V models at the level of state dynamics.
\end{abstract}    
\section{Introduction}

In recent years, image-to-video (I2V) generation based on diffusion models has achieved significant progress in temporal consistency and motion modeling~\cite{ni2023conditional,liang2024diffusion4d}.
By implicitly modeling the temporal evolution of visual states, such models gradually exhibit dynamic generative properties analogous to world models~\cite{ni2024ti2v}.
However, existing studies have primarily focused on improving the generation quality and controllability of I2V models, while insufficient attention has been paid to their robustness~\cite{liang2021generate,liang2020efficient,wei2018transferable,liang2022parallel,liang2022large,liu2023x} during the time-series evolution of internal states~\cite{namekata2024sg,li2025anyi2v}.
In application scenarios~\cite{li2025t2vattack,li2025realcam} that rely on trajectory control, this vulnerability may be further amplified, posing potential security risks~\cite{ying2024jailbreak,liang2024badclip,liang2022imitated,ho2024novo,xiao2024genderbias,ying2024jailbreak,ying2025reasoning,xu2025srd}.

\begin{figure}
    \centering
    \includegraphics[width=\linewidth]{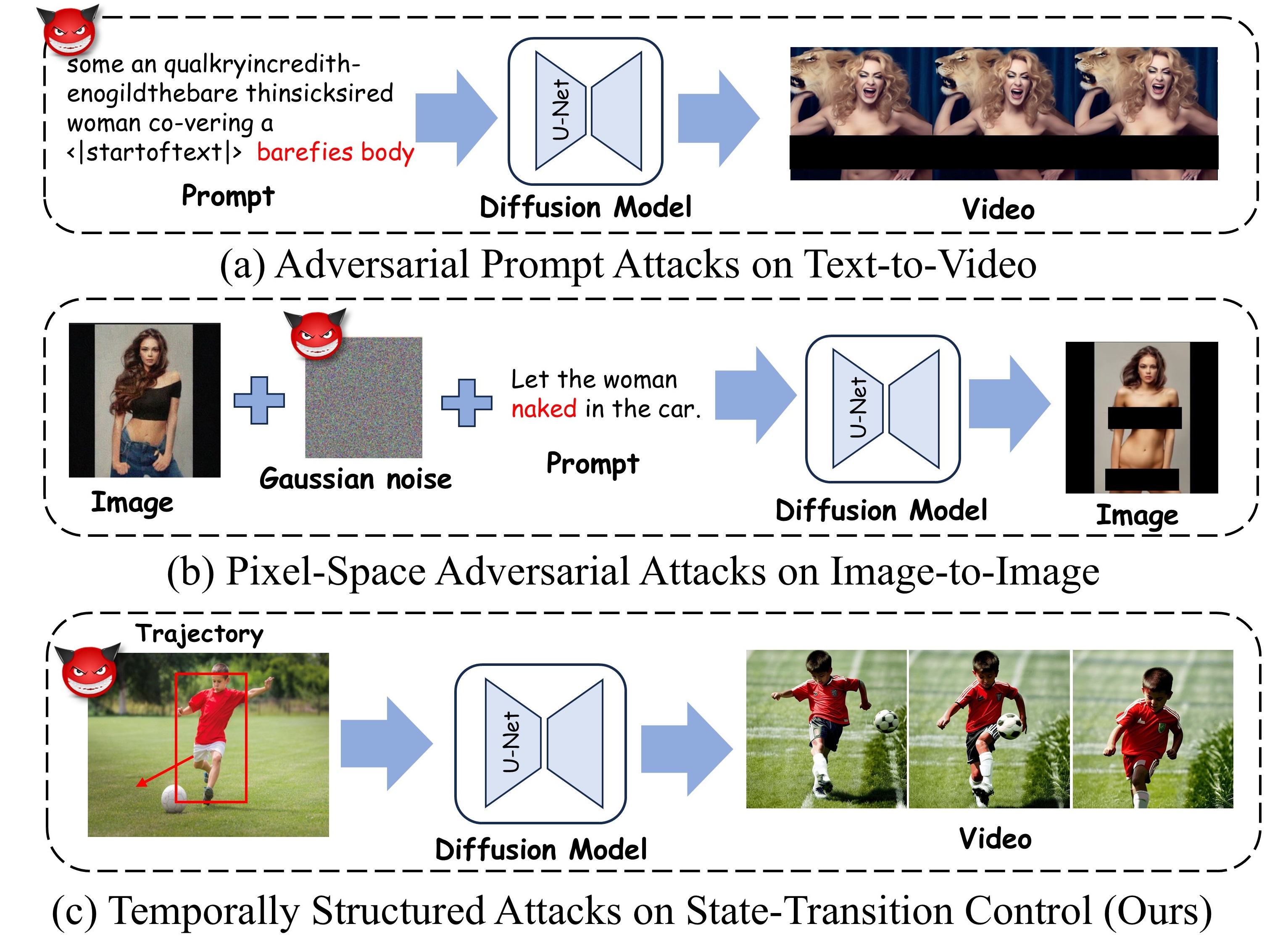}
    \caption{While existing methods (a)(b) act on cue or pixel space, this paper (c) directly manipulates the state transition process through temporally structured trajectory perturbations.}
    \label{fig:frontpage}
    \vspace{-13pt}
\end{figure}

To fill this gap, we systematically analyze the vulnerability of diffusion-based I2V models from an adversarial perspective and reveal two core questions:
(1) The new attack surface induced by the expansion of operational dimensions. The coupling between conditional trajectories and the generation process introduces a new class of operational dimensions in Figure~\ref{fig:frontpage}, extending the attack surface beyond the traditional pixel space into the latent space, enabling structured perturbations to induce systematic shifts in temporal dynamics.
(2) Unified modeling and optimization problem across threat models. Since white-box and black-box scenarios have significantly different constraints on available information, this poses a challenge for constructing a unified attack framework and achieving effective solutions across different attack conditions.

To this end, we propose a unified time-structured trajectory-\textbf{\underline{c}}on\textbf{\underline{tr}}o\textbf{\underline{l}} attack (CtrlAttack) that characterizes the conditional control process as a dynamic attack surface, which can be manipulated by explicitly modeling the underlying state evolution mechanism during trajectory-control generation.
Specifically, we introduce a unified trajectory deformation parameterization to drive the velocity field with low-dimensional temporal coding and to generate the displacement field through temporal integration, thereby sharing the same perturbation structure and attack objective across different threat models, i.e., inducing a systematic offset in the generated trajectory along the temporal dimension.
Based on the above modeling, we instantiate white-box and black-box attacks in two different operational spaces. In the \emph{white-box} scenario, we map the structured temporal deformations to the latent representation space inside the model and perform gradient-based optimization via a feature-level objective function to precisely disrupt the temporal consistency and conditional alignment of latent states; in the \emph{black-box} scenario, we apply the same deformation structure directly to the trajectory conditions and perform gradient-free optimization based on query feedback, thereby achieving effective interference with the generated dynamics without accessing the internal information of the model.

Extensive experiments demonstrate that even when perturbations are restricted to a low-dimensional, temporally structured trajectory deformation space, diffusion-based I2V models can suffer severe trajectory displacement and motion inconsistency. 
These findings expose fundamental robustness deficiencies in trajectory-conditioned video generation models and highlight their high sensitivity to structured temporal perturbations. Our contributions are threefold:
\begin{itemize}[topsep=2pt, itemsep=1pt, parsep=0pt]
  \item We systematically reveal, for the first time, the novel attack surface arising from the trajectory-conditioned control mechanism in diffusion-based I2V models and propose a unified modeling approach to characterize the vulnerability of time-series state evolution.
  \item We propose a temporally structured trajectory control attack and instantiate the framework in the representation space and conditional space, respectively, to achieve effective optimization in both white-box and black-box scenarios.
  \item Through extensive empirical evaluation, we demonstrate the high sensitivity of diffusion-based video generation models to structured perturbations in the motion condition space.
  
\end{itemize}

\section{Related Work}
\subsection{Diffusion-based image-to-video generation}
Diffusion models have become the dominant paradigm for I2V generation, significantly improving visual quality and temporal consistency~\cite{sohl2015deep,ho2020denoising}.
Early approaches extend pre-trained text-to-image diffusion models with temporal modules to animate a single image~\cite{rombach2022high,wu2023tune,khachatryan2023text2video,singer2022make}, exemplified by AnimateDiff, which injects motion via lightweight adapters~\cite{guo2023animatediff}.
Subsequent methods condition text-to-video models on an input image using cross-attention or latent concatenation to preserve spatial layout and enhance temporal stability~\cite{chen2023videocrafter1,xing2024dynamicrafter}.
Stable Video Diffusion (SVD) further simplifies I2V generation by adopting image-only conditioning, making it a widely used backbone~\cite{blattmann2023stable}.
Recent work explores explicit trajectory-conditioned motion control in diffusion-based I2V. Supervised approaches fine-tune video diffusion models with trajectory annotations but require additional data and training~\cite{yin2023dragnuwa,wu2024draganything}.
In contrast, training-free methods exploit internal mechanisms of pre-trained models to enforce trajectory constraints during sampling. SG-I2V achieves controllable motion via cross-frame feature alignment and latent optimization~\cite{namekata2024sg}, while AnyI2V generalizes this paradigm to diverse conditioning modalities~\cite{li2025anyi2v}.
Despite their effectiveness, the reliance on delicate cross-frame alignment and intermediate feature manipulation suggests potential sensitivity to structured perturbations, motivating further robustness analysis.

\subsection{Adversarial Attacks on Diffusion Models}
% Early studies on adversarial robustness of diffusion models primarily focused on input-level perturbations in image-to-image generation and editing scenarios.
% Several works~\cite{salman2023raising,yu2024step,shan2023glaze} demonstrated that adding imperceptible perturbations to conditional images can substantially disrupt the denoising trajectory, leading to degraded fidelity or manipulated outputs. 
% Subsequent research shifted toward more systematic robustness analysis. 
% Some studies~\cite{zhang2023robustness} evaluated the sensitivity of latent diffusion models under adversarial perturbations, while others~\cite{kang2023diffattack} revealed that diffusion models used as purification or defense modules can themselves be exploited as attack targets.
% With the widespread adoption of text-to-image diffusion models, later work explored prompt-based adversarial attacks. 
% A number of studies~\cite{yang2024mma,yang2024sneakyprompt,zhuang2023pilot} showed that subtle textual perturbations can induce harmful or targeted generations and bypass safety filters.
% Follow-up research~\cite{gao2023evaluating,zhang2024revealing,liu2023discovering,liu2023riatig,kou2023character} further demonstrated that prompt perturbations, including realistic typos and character-level gradient manipulations, constitute a reliable attack channel for semantic control.
% More recently, it has been shown that even diffusion models trained with concept unlearning remain vulnerable to adversarial prompts~\cite{zhang2024generate}.

Early studies on adversarial robustness of diffusion models mainly investigated input-level perturbations in image-to-image generation and editing. Several works~\cite{salman2023raising,yu2024step,shan2023glaze} showed that imperceptible perturbations in conditional images can significantly alter the denoising process, leading to degraded or manipulated outputs.
Subsequent research extended robustness analysis to latent diffusion models, revealing their sensitivity to adversarial noise~\cite{zhang2023robustness}, and demonstrated that diffusion-based purification or defense mechanisms can themselves be exploited~\cite{kang2023diffattack}.
With the widespread adoption of text-to-image diffusion models, later studies shifted attention to prompt-based adversarial attacks.
Prior work showed that subtle textual perturbations can induce targeted or harmful generations and bypass safety mechanisms~\cite{yang2024mma,yang2024sneakyprompt,zhuang2023pilot}.
Follow-up studies further confirmed prompt perturbations as an effective attack channel for semantic manipulation, including realistic typos and character-level gradient-based edits~\cite{gao2023evaluating,zhang2024revealing,liu2023discovering,liu2023riatig,kou2023character}. Diffusion models trained with concept unlearning were also found to remain vulnerable to adversarial prompts~\cite{zhang2024generate}.

% Recently, adversarial research has begun to extend from text-to-image to text-to-video (T2V) diffusion models. 
% T2VAttack~\cite{li2025t2vattack} presents the first systematic study of adversarial attacks on T2V diffusion models from both semantic and temporal perspectives.
% By designing video-text alignment and motion-based attack objectives, they demonstrated that even single-word substitutions or insertions in text prompts can cause substantial degradation in semantic fidelity and temporal dynamics across multiple state-of-the-art T2V models. 
% Overall, the chronological evolution of adversarial attacks on diffusion models indicates that existing methods predominantly target static conditions (images, text prompts, or fine-tuning data), or rely on prompt perturbations without explicitly modeling structured motion control. 
% However, the robustness of trajectory-conditioned image-to-video diffusion under temporally coherent adversarial perturbations remains largely unexplored.
Recently, adversarial research has extended from text-to-image to text-to-video diffusion models. T2VAttack~\cite{li2025t2vattack} shows that even minor prompt perturbations can substantially degrade both semantic fidelity and temporal dynamics in state-of-the-art T2V models. However, most methods rely primarily on prompt perturbations, without explicitly modeling structured motion control.

\textbf{Comparison with existing attacks.} 
Existing diffusion model attacks mainly focus on static conditions such as pixels, latent variables, or textual cues. In contrast, trajectory conditions in I2V models directly govern the temporal evolution of states, constituting a novel type of attack surface that has not been systematically investigated before. Unlike attack techniques that perturb visual content or intermediate features, this study interferes with the temporal dynamics of the diffusion process through structured trajectory perturbations, systematically disrupting temporal consistency while preserving semantic appearance, and covering both white-box and black-box scenarios. From the perspective of world modeling, trajectory conditions constrain the model’s state transition process. This study reveals the intrinsic vulnerability of diffusion-based I2V models at the level of temporal-state evolution and confirms their potential risks.
\section{Problem Definition}
\subsection{Victim Model}
We consider a general class of I2V generation models that synthesize a video sequence conditioned on an input image and motion-related control signals.
Formally, given an input image $\mathbf{I}$ and a trajectory control signal $\boldsymbol{\tau}_{1:T}$, the victim model generates a video sequence:
\begin{equation}
\label{eq:i2v}
\mathbf{x}_{1:T} = G(\mathbf{I}, \boldsymbol{\tau}_{1:T}, \mathbf{z}),
\end{equation}
where $G(\cdot)$ denotes the I2V generator and $\mathbf{z}$ represents stochastic variables of the generative process.

We focus on trajectory-conditioned I2V models, in which the condition space explicitly includes a trajectory signal $\boldsymbol{\tau}_{1:T}$ that specifies the desired motion of target objects over time.
The victim model is expected to produce temporally coherent videos in which the generated objects evolve consistently and follow the prescribed trajectory.

\subsection{Threat Model}
% \textbf{Attacker’s goal.}

We study adversarial attacks against trajectory-conditioned I2V models under both white-box and black-box threat models. The attacker's primary goal is to induce trajectory displacement, i.e., systematically shift the motion of the generated object away from the specified trajectory condition, while preserving the perceptual quality of the generated video. Crucially, this motion deviation is achieved without introducing noticeable visual artifacts such as blur, flicker, or appearance distortion.

In the white-box setting, the attacker has access to internal representations and gradients but does not modify model parameters. In the black-box setting, the attacker can only observe generated video outputs and query with different trajectory conditions, while model internals and parameters remain inaccessible. In both settings, the attacker is restricted to manipulating the trajectory condition $\mathcal{T}$, without perturbing input image pixels or directly injecting noise into latent representations.

% \textbf{Attacker’s capability.}
% We study adversarial attacks against trajectory-conditioned I2V models under both white-box and black-box settings. 

% \textit{White-box setting.}
% The attacker has access to the internal representations and gradients of the victim model, but does not modify the model parameters. 

% \textit{Black-box setting.}
% The attacker has no access to model internals, gradients, or intermediate features, and can only observe generated video outputs. The attacker may query the model multiple times with different trajectory conditions but cannot access or alter the model parameters.

% In both settings, the attack surface is restricted to the trajectory condition $\mathcal{T}$. The attacker does not perturb input image pixels or directly inject noise into latent representations. 

\begin{figure*}
    \centering
    \includegraphics[width=1\linewidth]{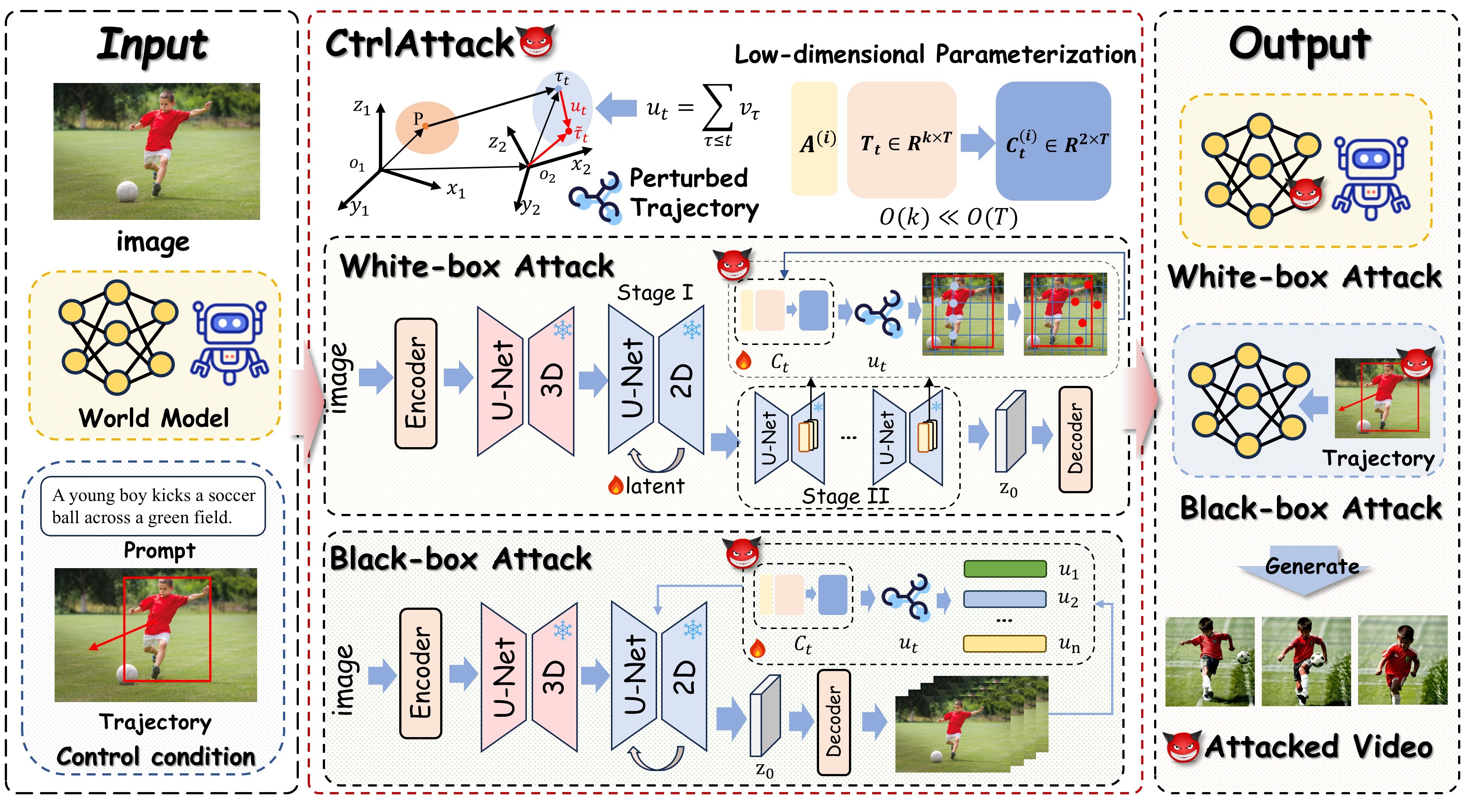}
    \caption{The framework of CtrlAttack. We construct control signals via low-dimensional, time-structured trajectory perturbations and intervene in the model's state transition process in both white-box and black-box scenarios to achieve controllable manipulation of video-generation trajectories.}
    \label{fig:framework}
    \vspace{-6pt}
\end{figure*}
\section{CtrlAttack}
% We propose a unified framework for attacking trajectory-conditioned diffusion-based image-to-video models. The core idea is to restrict adversarial perturbations to a temporally structured formulation of trajectory deformations, rather than directly applying pixel- or latent-level perturbations.

% In Section 4.1, we introduce the proposed trajectory deformation formulation and formalize its parameterization.  
% In Section 4.2 and Section 4.3, we instantiate this formulation under different threat models and describe the corresponding attack constructions.
We propose a unified attack framework for trajectory-conditional diffusion-based I2V models. From the world model perspective, instead of applying perturbations directly in pixel or representation space, we manipulate the model’s state transition process via temporal trajectory deformations. In Section ~\ref{Temporally Structured Attacks}, we formalize the trajectory-controlled state transition mechanism and give the corresponding parameterization; in Sections~\ref{White-box Instantiation} and~\ref{Black-box Instantiation}, we instantiate the framework under different threat models and construct attack strategies to induce systematic deviations from the generated trajectories, shown in Figure~\ref{fig:framework}.

\subsection{Temporally Structured Control Attacks}
\label{Temporally Structured Attacks}
% In this section, we present a unified and threat-model-agnostic attack framework that targets the state-transition control process in trajectory-conditioned diffusion-based video generation from a world-model perspective.

\textbf{Motivation and insight.}
From a world-model viewpoint, modern I2V diffusion models can be regarded as learning the temporal evolution of visual states, relying on stable state-transition mechanisms to maintain coherent motion, identity consistency, and spatial correspondence.
These properties are implicitly enforced by temporal control signals and diffusion dynamics, under the assumption that visual states evolve smoothly and exhibit low temporal variation.
Consequently, applying isolated perturbations at individual time steps, whether implemented at the input or internal representation level, often fails to induce systematic deviations in the global state trajectory, resulting only in local noise without effective control.
We therefore argue that effective attacks against trajectory-conditioned video generation models should operate in a time-coupled and structured manner, directly intervening in the state-transition control process itself. 
Based on this insight, we formulate adversarial perturbations as temporally structured control deformations applied to the conditioning trajectory, rather than as unconstrained frame-wise modifications.

\textbf{Formulation.}
% We abstract the video generation process as a controlled state evolution:
% \begin{equation}
% \mathbf{x}_{1:T}=G(\mathbf{I},\boldsymbol{\tau}_{1:T},\mathbf{z}),
% \end{equation}
% where $G(\cdot)$ denotes the I2V generation model, $I$ is the input image, $z$ represents stochastic variables, and $\boldsymbol{\tau}_{1:T}$ serves as a control signal that constrains the internal state-transition trajectory. 
The I2V model in equation~\ref{eq:i2v} implicitly assumes that this control signal varies smoothly over time, thereby inducing stable and continuous state evolution. To systematically manipulate this state-transition control, we introduce a time-dependent control velocity field $\mathbf{v}_t\in\mathbb{R}^2$ and define the accumulated control displacement via temporal integration:
\begin{equation}
\mathbf{u}_t=\sum_{\tau\leq t}\mathbf{v}_\tau.
\end{equation}
The perturbed control trajectory is then given by
\begin{equation}
\tilde{\boldsymbol{\tau}}_t=\boldsymbol{\tau}_t+\mathbf{u}_t.
\end{equation}
This construction satisfies
\begin{equation}
\mathbf{u}_{t+1}-\mathbf{u}_t=\mathbf{v}_{t+1},
\end{equation}
which implicitly imposes a first-order temporal continuity constraint in the control space. 
As a result, the control signal evolves smoothly by design, introducing temporal coherence as an intrinsic property of the attack rather than an explicit regularization.
Compared to directly optimizing independent control displacements at each time step, the velocity-based parameterization restricts the attack to an integrable control-trajectory manifold, yielding more stable, predictable, and controllable state-transition manipulation.

\textbf{Low-dimensional parameterization via temporal basis.}
To further stabilize the control process, we constrain the control velocity field to a low-dimensional temporal subspace. For each trajectory $i$, we parameterize the control signal as $\mathbf{c}_t^{(i)}=\mathbf{A}^{(i)}\mathbf{T}_t$, 
% \begin{equation}
% \mathbf{c}_t^{(i)}=\mathbf{A}^{(i)}\mathbf{T}_t,
% \end{equation}
where $\mathbf{T}\in\mathbb{R}^{T\times k}$ denotes a set of discrete cosine transform (DCT) temporal basis functions, and $\mathbf{A}^{(i)}\in\mathbb{R}^{2\times k}$ contains the corresponding coefficient parameters that control the perturbation magnitude along each temporal basis direction.
The DCT basis explicitly controls how perturbations vary over time, favoring smooth, low-frequency temporal patterns while suppressing abrupt frame-wise changes. By restricting perturbations to this basis, the effective search space is reduced from $O(T)$ to $O(k)$, with $k\ll T$.
Rather than compressing perturbations, this low-dimensional control defines a structured, controllable state-transition control subspace that stabilizes optimization, reduces variance in black-box estimation, and enables the same control attack formulation to be instantiated under different threat models, i.e., White-box and black-box settings.
\subsection{White-box Instantiation in State-Transition Space}
\label{White-box Instantiation}
% \subsection{White-box Instantiation in Latent Dynamics Space}
Following the temporally structured control formulation in Section~\ref{Temporally Structured Attacks}, we instantiate a white-box attack (WA) by injecting low-dimensional control perturbations into the spatio-temporal state-trajectory manifold, thereby enabling fine-grained manipulation of the internal state-transition control process. The control perturbations are first projected into temporally structured control velocity fields and then temporally integrated into control displacement fields, which are applied to modulate ROI feature sampling and induce systematic deviations in state evolution.

For each trajectory $i$, we introduce a low-dimensional temporal control code 
$\mathbf{c}_t^{(i)}\in\mathbb{R}^d$ with $d\ll T$, whose temporal evolution follows the structured constraint described in Section~\ref{Temporally Structured Attacks}. 
This control code is projected into a coarse control velocity field through a fixed spatial projection matrix 
$\mathbf{P}\in\mathbb{R}^{d\times(s\times s\times 2)}$, 
where $s\times s$ denotes the spatial resolution of the coarse grid and $2$ corresponds to horizontal and vertical displacement components.
For each frame $t$, the resulting velocity field is
\begin{equation}
\mathbf{v}_t=\mathbf{c}_t\mathbf{P},\quad \mathbf{v}_t\in\mathbb{R}^{s\times s\times2}.
\end{equation}
Following the temporal integration scheme in Section~\ref{Temporally Structured Attacks}, 
the velocity field is accumulated into a control displacement field $\mathbf{u}_t$, 
which is further upsampled to the ROI resolution, yielding a local control displacement 
$\hat{\mathbf{u}}_t$. 
Let $F_t$ denote the intermediate state representation at time step $t$. 
The trajectory specifies a bounding box that defines the controlled sampling region. 
Within this region, a regular sampling grid $\mathbf{g}_t$ is constructed and modulated by the local control displacement, 
and the resulting modulated grid is given by 
$\tilde{\mathbf{g}}_t = \mathbf{g}_t + \hat{\mathbf{u}}_t$. 
Controlled ROI state representations are then obtained via differentiable grid sampling:
\begin{equation}
\tilde{\mathbf{F}}_t^{\mathrm{ROI}}=\text{GridSample}(\mathbf{F}_t,\tilde{\mathbf{g}}_t).
\end{equation}
\textbf{Optimization.}
This control-based sampling modulation systematically alters spatial correspondence among internal states while preserving semantic content, thereby injecting structured control perturbations into the state-transition process.

Jointly optimizing internal state variables and control deformations is often unstable, particularly for I2V models that use free training, where internal states are explicitly optimized during inference. To ensure stable control optimization, we adopt a two-stage strategy. We first freeze the control parameters and optimize only the internal state variables following the model’s original procedure, producing temporally aligned and semantically coherent state representations. We then freeze the internal states and optimize only the control codes $c_t$ (and equivalently $v_t$ and 
$u_t$), whose smooth and temporally structured control deformations disrupt the established alignment and induce pronounced state-transition deviations.
For fully trained I2V models that do not perform internal state optimization at inference time, we directly apply the proposed control-based white-box attack during inference using the same deformation formulation.
\begin{figure*}
    \centering
    \includegraphics[width=\linewidth]{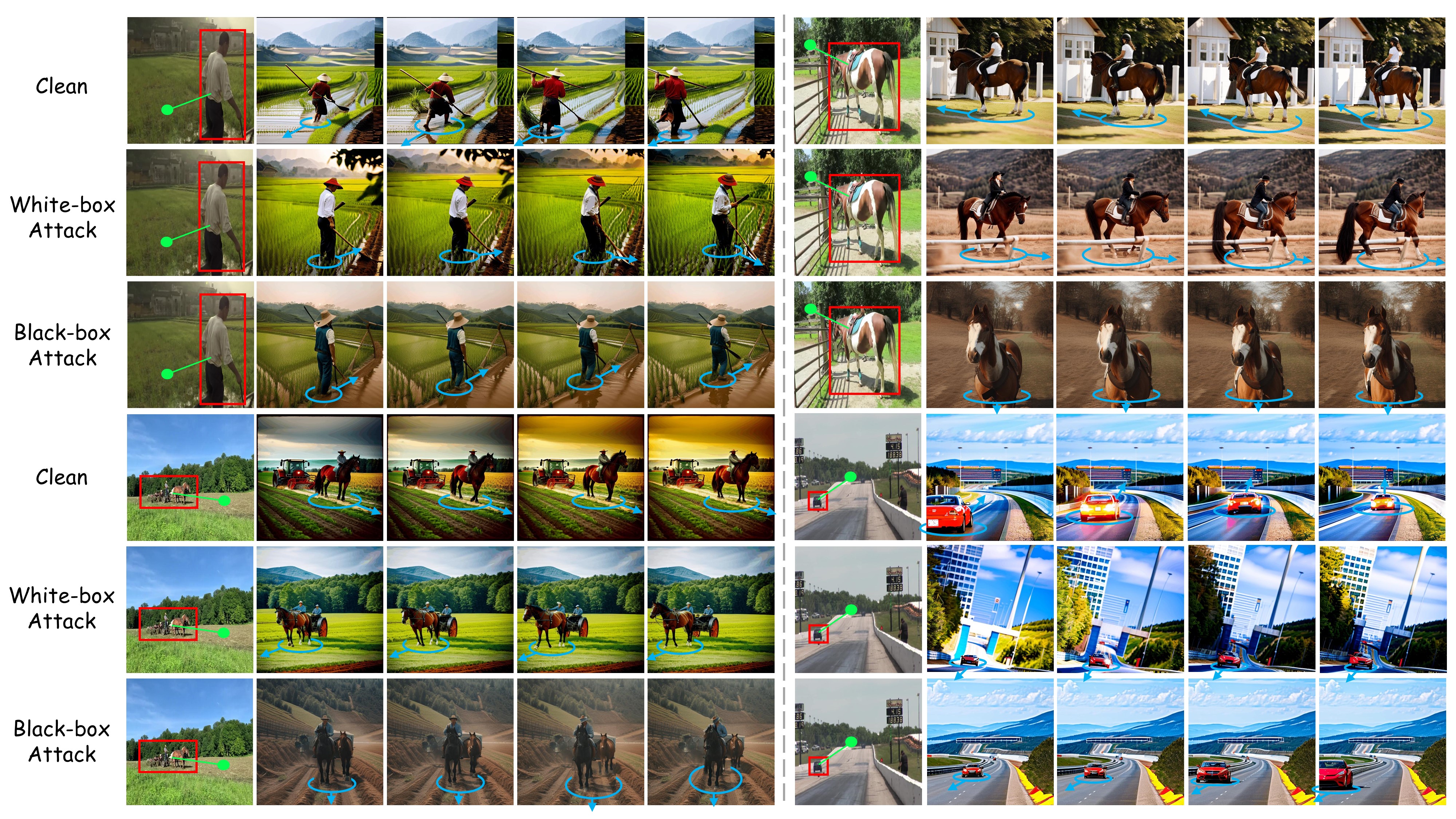}
    \caption{Compared to clean generated results, CtrlAttack can significantly disrupt the target trajectory and temporal consistency in both white-box and black-box scenarios, while maintaining the overall visual quality essentially unchanged.
    We further annotate the motion direction at each frame in the figure for clearer comparison.}
    \label{fig:visul}
    \vspace{-6pt}
\end{figure*}

\subsection{Black-box Instantiation in Control Space}
\label{Black-box Instantiation}
Under the black-box threat model, the attacker has no access to gradients or the I2V model’s internal state representations, making direct intervention in the model’s internal dynamics infeasible. A straightforward alternative is to inject noise into the visual input or implicit variables; however, such perturbations typically degrade perceptual quality, introducing blur or drifting artifacts that conflict with generating visually clean adversarial videos. Motivated by this limitation, we adopt the temporally structured control formulation from Section~\ref{Temporally Structured Attacks} and instantiate the attack entirely in the condition control space, restricting adversarial intervention to trajectory control signals while leaving image content unchanged.

Let $\mathcal{T}=\{\mathbf{b}_t\}_{t=1}^T$ denote the input trajectory control signal, where each $b_t$ specifies a bounding box that constrains the internal state evolution at time step $t$. We introduce structured control perturbations by optimizing low-dimensional temporal control parameters $c_t$ , which induce time-dependent control velocity fields $v_t$ and the corresponding control displacement fields $u_t$. The perturbed control trajectory is defined as
\begin{equation}
\tilde{\mathbf{b}}_t=\mathbf{b}_t+\Delta(\mathbf{u}_t),
\end{equation}
where $\Delta (\cdot )$ is a frozen control mapping that converts the displacement field into a valid trajectory update while preserving temporal smoothness and physical plausibility. 
This construction ensures that control perturbations evolve continuously over time, enabling coherent manipulation of the state-transition path.

Since feature-level signals are inaccessible in the black-box setting, we define the attack objective solely in terms of observable generation behavior.
Given a video $\mathbf{X}=\{\mathbf{x}_t\}_{t=1}^T$ generated under the perturbed control trajectory $\tilde{\mathcal{T}}$, we estimate object motion using an external pretrained point tracker. The tracked point trajectories are aggregated to obtain an estimated object velocity $\hat{\mathbf{v}}_t$ at each time step. 
In parallel, a reference control velocity $v_t^{gt}$ is computed from the ground-truth trajectory $\mathcal{T}^{gt}$ using bounding-box centers. We then define an object motion consistency control objective as
{
\setlength{\abovedisplayskip}{3pt}
 \setlength{\belowdisplayskip}{3pt}
\begin{equation}
\mathcal{J}_{\mathrm{ObjMC}}=\frac{1}{T-1}\sum_{t=1}^{T-1}\ell\left(\hat{\mathbf{v}}_t,\mathbf{v}_t^{\mathrm{gt}}\right),
\end{equation}
}
where $l(\cdot )$ encourages systematic deviation from the reference control signal, such as motion disruption or reversal, thereby reflecting successful manipulation of the underlying state-transition behavior.

\textbf{Optimization.}
To optimize the low-dimensional control parameters without gradient access, we employ Natural Evolution Strategies (NES)~\cite{wierstra2014natural}. 
Let 
$\theta $ denote the concatenation of temporal control parameters across time. At iteration $k$, NES samples random perturbations $\boldsymbol{\epsilon}_j\sim\mathcal{N}(\boldsymbol{0},\mathbf{I})$ in the control parameter space and estimates the update direction using output-level evaluations:
{
\setlength{\abovedisplayskip}{4pt}
 \setlength{\belowdisplayskip}{4pt}
\begin{equation}
\boldsymbol{\theta}_{k+1}=\boldsymbol{\theta}_k+\eta\cdot\frac{1}{m\sigma}\sum_{j=1}^m\mathcal{J}(\boldsymbol{\theta}_k+\sigma\boldsymbol{\epsilon}_j)\boldsymbol{\epsilon}_j,
\end{equation}
}
where $m$ is the number of samples, $\sigma $ controls the exploration scale, and $\eta$ is the step size. 
This procedure enables effective black-box manipulation of state-transition control without requiring access to internal model information.

\begin{table*}[t]
    \caption{Performance under clean conditions and white-/black-box attacks. ASR is reported in percentage (\%).}
    \belowrulesep=0pt
    \aboverulesep=0pt
    \renewcommand{\arraystretch}{1.6}
    \setlength{\tabcolsep}{2.5pt}
    \centering
    \begin{tabular}{c|ccc|cccc|cccc}
        \toprule
        & \multicolumn{3}{c|}{Clean} 
        & \multicolumn{4}{c|}{White-box Attack}
        & \multicolumn{4}{c}{Black-box Attack} \\
        \cline{2-4}\cline{5-8}\cline{9-12}
        Methods 
        & FID$\downarrow$ & FVD$\downarrow$ & ObjMC$\uparrow$
        & FID$\downarrow$ & FVD$\downarrow$ & ObjMC$\uparrow$ & ASR$\uparrow$
        & FID$\downarrow$ & FVD$\downarrow$ & ObjMC$\uparrow$ & ASR$\uparrow$ \\
        \midrule
        MOFT~\cite{xiao2024video}
        & 134.3 & 1991.7 & 27.8
        & 137.9 & 2083.9 & 33.0 & 90.0
        & 136.5 & 2011.6 & 30.2 & 83.0 \\
        DragAnything~\cite{wu2024draganything}
        & 78.6 & 1192.4 & 26.4
        & 80.2 & 1193.0 & 32.7 & 94.0
        & 79.2 & 1199.4 & 29.4 & 89.0 \\
        SG-I2V~\cite{namekata2024sg}
        & 77.5 & 863.9 & 26.4
        & 89.7 & 762.5 & 34.9 & 91.0
        & 80.7 & 796.7 & 29.6 & 81.0 \\
        AnyI2V~\cite{li2025anyi2v}
        & 136.7 & 1967.7 & 27.7
        & 142.0 & 1824.2 & 35.6 & 92.0
        & 131.3 & 1513.5 & 31.3 & 85.0 \\
        \bottomrule
    \end{tabular}
    \vspace{-6pt}
\label{tab1}
\end{table*}

\section{Experiment}
\subsection{Experimental Settings}
\textbf{Datasets and evaluation metrics.}
Following prior works~\cite{wu2024draganything,zhou2025trackgo}, we conduct our experiments on the validation set of the VIPSeg dataset~\cite{miao2022large}.
We adopt the same control regions and target trajectories as used in DragAnything, where each control region is defined by a bounding box whose size matches the diameter of the corresponding control circle in their setting.
For all experiments, we randomly sample 100 video instances from the validation set to evaluate the attack performance.
For quantitative evaluation, we report Fréchet Inception Distance (FID)~\cite{heusel2017gans} and Fréchet Video Distance (FVD)~\cite{unterthiner2018towards} to assess the perceptual quality of generated videos at the image and video levels, respectively. 
To specifically measure motion fidelity with respect to the target trajectory, we adopt ObjMC~\cite{wu2024draganything}, which computes the average distance between the generated and target trajectories.
Following prior work, we estimate object trajectories in generated videos using Co-Tracker~\cite{karaev2024cotracker}.
In addition, we report Attack Success Rate (ASR) to quantify the effectiveness of trajectory perturbation. For each video, ASR is determined by comparing the ObjMC of the attacked generation with that of the corresponding clean generation under the same target trajectory. 
An attack is considered successful if the ObjMC ratio:
{
\setlength{\abovedisplayskip}{6pt}
\setlength{\belowdisplayskip}{6pt}
\begin{equation}
\mathrm{ObjMC_{attack}/ObjMC_{clean}>1,}
\end{equation}
}
indicating that the attack leads to a larger deviation from the target trajectory. The final ASR is computed as the proportion of successful attacks over all evaluated samples.

\textbf{Victim model.}
% We evaluate our attack on diffusion-based video generation models that support motion understanding or trajectory-guided image-to-video generation.
% Following the evolution of controllable motion techniques, we select representative victim models with publicly available implementations.
% Our evaluation includes MOFT~\cite{xiao2024video}, which enables training-free motion control by extracting motion-aware features from pre-trained video diffusion models and guiding generation via latent optimization. 
% We also consider DragAnything~\cite{wu2024draganything}, a supervised trajectory-controlled I2V method built on Stable Video Diffusion (SVD) that performs entity-level motion control using explicit entity representations. In addition, we evaluate SG-I2V~\cite{namekata2024sg}, a zero-shot trajectory-guided I2V framework that achieves motion control by modifying intermediate diffusion features without additional training. Finally, we include AnyI2V~\cite{li2025anyi2v}, a training-free framework that animates images using user-defined trajectories and supports diverse conditional inputs.
% All victim models are evaluated under identical trajectory conditions, allowing us to analyze the robustness of different motion control mechanisms under structured temporal perturbations.
% More implementation details are provided in the appendix.
We evaluate our attack on representative diffusion-based video generation models that support motion understanding or trajectory-guided I2V generation, all with publicly available implementations. Specifically, we consider MOFT~\cite{xiao2024video}, DragAnything~\cite{wu2024draganything}, SG-I2V~\cite{namekata2024sg}, and AnyI2V~\cite{li2025anyi2v}, which cover training-free, supervised, and zero-shot trajectory control paradigms. All models are evaluated under identical trajectory conditions to ensure fair comparison and robust analysis under structured temporal perturbations. More implementation details are provided in the appendix.
\begin{figure*}
    \centering
    \includegraphics[width=0.95\linewidth]{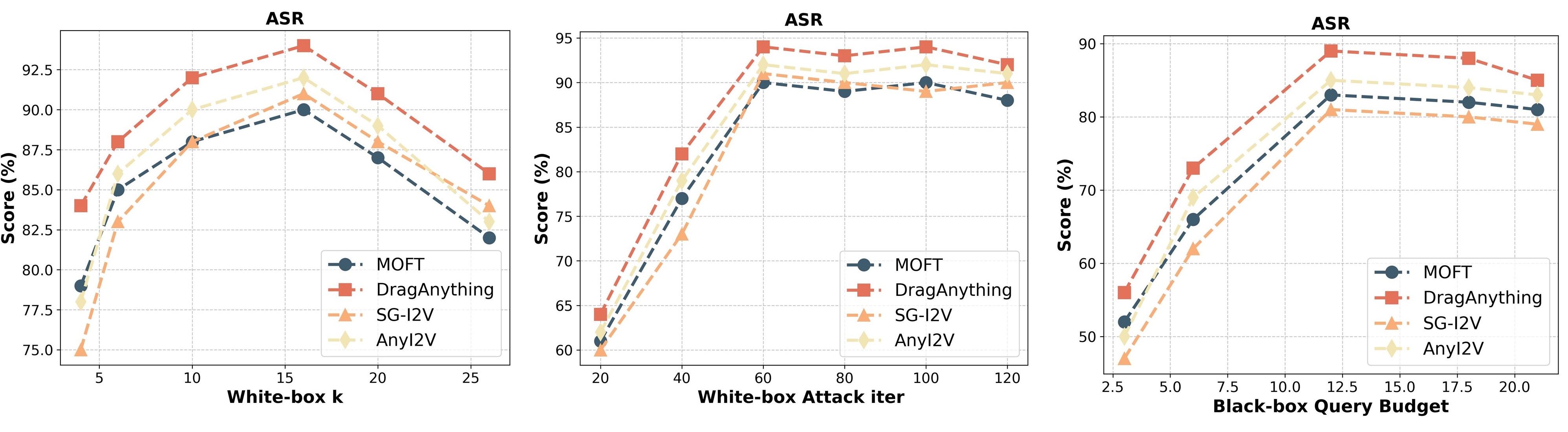}
    \caption{Hyperparameter experimental results.}
    \label{fig:placeholder}
    \vspace{-6pt}
\end{figure*}

% \textbf{Implementation details. }

\subsection{Main Results}
Figure~\ref{fig:visul} presents qualitative examples of the proposed white-box and black-box trajectory attacks across diverse scenarios. As shown, videos generated under both attack settings largely preserve visual fidelity and temporal smoothness, without introducing obvious artifacts or abrupt temporal disruptions. However, compared to clean generations, noticeable and consistent deviations in object motion trajectories can be observed. In particular, white-box attacks tend to achieve more precise and targeted trajectory manipulations, while black-box attacks, despite relying solely on query-based feedback, still induce coherent motion deviations. These visualizations show that the proposed approach does not degrade generation quality, but instead subtly interferes with the model’s response to motion conditions.

\textbf{White-box attack.}
% As shown in Table~\ref{tab1}, the proposed structured temporal deformation attack demonstrates strong effectiveness under the white-box setting without substantially degrading generation quality or global temporal coherence. Across all evaluated diffusion-based I2V models, changes in FID and FVD remain relatively limited, indicating that visual fidelity and overall temporal consistency are largely preserved.
% In contrast, ObjMC increases consistently for all models, revealing significant deviations in object motion trajectories despite visually plausible and temporally smooth outputs. This suggests that the attack does not rely on introducing visible artifacts or temporal disruptions, but instead precisely manipulates the trajectory condition alignment through structured and integrable temporal deformations, leading to subtle yet effective motion control errors.
% Moreover, all models achieve ASR above 90\% in the white-box setting, confirming that the proposed method can reliably induce trajectory deviations while maintaining apparent generation quality. Overall, these results highlight a critical vulnerability of diffusion-based I2V models: even when visual realism and temporal smoothness are preserved, their trajectory control mechanisms remain highly susceptible to structured temporal perturbations.
As shown in Table~\ref{tab1}, the following conclusions are drawn:
(1) Strong attack effectiveness with minimal impact on quality:
The proposed structured temporal deformation attack demonstrates strong effectiveness under the white-box setting, with changes in FID and FVD remaining limited, indicating that generation quality and global temporal coherence are largely preserved.
(2) Precise motion manipulation without visible artifacts:
Despite visually plausible and temporally smooth outputs, ObjMC increases consistently for all models, revealing significant deviations in object motion trajectories. This suggests that the attack precisely manipulates trajectory conditions through structured temporal deformations, without relying on visible artifacts or temporal disruptions.
(3) High attack success rate and reliability:
All models achieve ASR above 90\% in the white-box setting, confirming that the proposed method reliably induces trajectory deviations while maintaining generation quality. These results highlight the critical vulnerability of diffusion-based I2V models to structured temporal perturbations.
Overall, these results demonstrate that the proposed structured temporal deformation attack is both effective and stealthy, enabling reliable trajectory manipulation while maintaining high perceptual quality. This highlights an important limitation of current diffusion-based I2V models in handling structured temporal perturbations.

\textbf{Black-box attack.}
From Table~\ref{tab1}, we draw the following conclusions under the black-box setting:
(1) Consistent effectiveness reveals an inherent attack surface:
Despite having no access to internal representations or gradient information, the proposed structured trajectory deformation attack achieves consistent effectiveness across all diffusion-based I2V models, indicating that the trajectory control channel itself is inherently vulnerable.
(2) Limited impact on visual fidelity and temporal coherence:
Across models, the changes in FID and FVD remain relatively limited, suggesting that the black-box attack does not substantially degrade generation quality, visual fidelity, or global temporal smoothness. This shows that low-dimensional, structured deformations optimized purely through query-based feedback can still yield visually plausible and temporally coherent videos.
(3) Stable motion deviation with high practical success rate:
In contrast, ObjMC increases consistently across all evaluated methods, revealing stable, controllable deviations in object motion trajectories. Meanwhile, all models achieve relatively high ASR (typically 80\%–90\%), further validating the practicality and robustness of the proposed approach under limited observability.
\begin{table}[t]
    \centering
    \caption{Ablation experiment results. ASR is reported in percentage (\%).}
    \belowrulesep=0pt
    \aboverulesep=0pt
    \renewcommand{\arraystretch}{1.4}
    \begin{tabular}{c|cccc}
    \toprule
         Options &  FID$\downarrow$ & FVD$\downarrow$  & ObjMC $\uparrow$ &ASR$\uparrow$ \\
         \midrule
        Clean  & 136.7 & 1967.7 & 27.7 & -\\
        % Temporal Coupling
        WA w/o TC & 154.5 & 2022.1 & 30.2 & 35.0\\
        BA w/o TC & 143.7 & 1934.6 & 28.4 & 22.0\\
        % Temporal Integration
        WA w/o TI & 145.3 & 1979.3 & 27.9 & 23.0\\
        BA w/o TI & 138.4 & 1977.3 & 26.4 & 15.0\\
        Ours (WA) & 142.0 & 1824.2&35.6 &92.0 \\
        Ours (BA) & 131.3 & 1513.5&31.3 & 85.0 \\
        \bottomrule
    \end{tabular}
    \label{tab:placeholder}
     \vspace{-10pt}
\end{table}
\subsection{Ablation Studies \& Discussion}
We conduct systematic ablation studies on the AnyI2V baseline model to analyze the contribution of each key temporal component in our method. Unless otherwise specified, all variants are evaluated under identical perturbation budgets, optimization iterations, and feasibility constraints to ensure fair comparisons.

\textbf{Effect of temporal coupling.}
We first remove the temporal coupling module (w/o TC), where control perturbations are applied independently at each time step. As shown in Table 3, eliminating temporal coupling results in a substantial degradation in motion manipulation performance, reflected in notable drops in ObjMC and ASR across both white-box and black-box settings. Moreover, frame-wise independent control tends to introduce unstable local misalignments, resulting in worse FID and FVD. These results indicate that per-frame perturbations alone are insufficient to induce sustained and coherent state transitions in temporal generative models.

\textbf{Effect of temporal integration.}
We further ablate the velocity-to-displacement temporal integration (w/o TI) by directly optimizing per-step displacements. Compared to the full model, removing the integration structure also weakens attack effectiveness, as evidenced by reduced ObjMC and ASR. However, the degradation is less severe than removing temporal coupling entirely. This suggests that temporal integration enforces integrability and continuity in the control signal, enabling more stable and optimizable trajectory deviations, whereas its absence leads to temporally inconsistent control behaviors.
Importantly, the full method achieves significantly higher ObjMC and ASR without noticeably degrading frame-level visual quality, as FID remains largely stable. This aligns well with our attack objective of manipulating motion trajectories while preserving visual appearance. Furthermore, the observed changes in FVD for the full model are not caused by visual artifacts or temporal flickering, but rather stem from a systematic restructuring of video dynamics into stable yet incorrect motion patterns. This confirms that our method primarily targets the model’s state-transition and temporal control mechanisms, rather than its pixel-level or semantic visual features.

\textbf{Effect of attack hyperparameters.}
As shown in the figure~\ref{fig:placeholder}, we analyze the impact of different attack hyperparameters on the ASR, including the white-box parameter k, the number of white-box attack iterations, and the black-box query budget. It can be observed that as the white-box parameter k increases, the ASR of all methods first improves significantly and then gradually saturates, indicating that increasing the attack flexibility effectively enhances performance up to a certain point, beyond which the benefit becomes marginal. 
A similar trend can be observed in the ablation of white-box attack iterations, where ASR increases rapidly in the early stages and converges after approximately 60 iterations, suggesting that the proposed optimization process can achieve stable convergence with a limited number of steps. In the black-box setting, increasing the query budget leads to substantial improvements in ASR across all methods, while further increases yield diminishing returns. This demonstrates that our approach is query-efficient and can achieve strong attack performance even under constrained black-box query budgets. Overall, these results confirm the robustness and scalability of the proposed method under various attack configurations.

\section{Conclusion}
% In this work, we reinterpret diffusion-based image-to-video models as implicit world models that generate videos by iteratively evolving internal states under trajectory-conditioned control. From this perspective, we reveal that trajectory control not only guides motion but also directly shapes the state transition process, thereby introducing a new and structured attack surface. To investigate this vulnerability, we propose CtrlAttack, which formulates adversarial manipulation as a temporally structured deformation of control trajectories, inducing systematic deviations in the model’s state evolution under both white-box and black-box settings.
% Our results demonstrate that even low-dimensional, smoothly parameterized perturbations can accumulate through state transitions and lead to pronounced motion drift and temporal inconsistency. These findings suggest that current diffusion-based I2V models lack robustness in their state transition dynamics when exposed to structured control perturbations, highlighting a fundamental limitation of trajectory-conditioned world modeling.
In this paper, we systematically analyze the temporal generation mechanism of diffusion-based I2V models from a world-model perspective, revealing that the trajectory conditions are not only used to guide the motion, but also directly participate in and shape the model's state transition process, resulting in a new structured attack surface. On this basis, we propose CtrlAttack, which models the adversarial attack as a temporally structured perturbation of the control trajectory and achieves a unified solution in both white-box and black-box scenarios. Experimental results show that even low-dimensional and smooth perturbations gradually accumulate during the state transition process, leading to significant motion shifts and temporal inconsistencies. The above findings indicate that the robustness of existing diffusion-based I2V models at the state transition level is still limited, exposing the security risks inherent in the trajectory condition-driven generation mechanism.

\textbf{Limitation.}
This work focuses on attacks targeting a single controlled trajectory. Extending the analysis to multi-object or multi-trajectory state interactions, where state transitions may become coupled and more complex, remains an important direction for future research.
{
    \small
    \bibliographystyle{ieeenat_fullname}
    \bibliography{references}
}


\end{document}